\def\year{2020}\relax
\documentclass[letterpaper]{article} 
\usepackage{aaai20}  
\usepackage{times}  
\usepackage{helvet} 
\usepackage{courier}  
\usepackage[hyphens]{url}  
\usepackage{graphicx} 
\usepackage{graphicx}  
\usepackage{amsmath}
\usepackage{bbm}
\usepackage{subcaption}
\usepackage{amssymb}

\DeclareMathOperator*{\argmin}{argmin}
\title{Morphing and Sampling Network for Dense Point Cloud Completion\thanks{We would like to thank Yen-Hsiu Chou and Zhengping Zhou for helpful comments.}}
\author{Minghua Liu,\textsuperscript{\rm 1} Lu Sheng,\textsuperscript{\rm 2} Sheng Yang,\textsuperscript{\rm 3} Jing Shao,\textsuperscript{\rm 4} Shi-Min Hu\textsuperscript{\rm 3} \\
\textsuperscript{\rm 1}UC San Diego, 
\textsuperscript{\rm 2}Beihang University, 
\textsuperscript{\rm 3}Tsinghua University, 
\textsuperscript{\rm 4}Sensetime \\
minghua@ucsd.edu, lsheng@buaa.edu.cn, shengyang93fs@gmail.com,\\ shaojing@sensetime.com, shimin@tsinghua.edu.cn
}
 
\begin{document}

\maketitle

\begin{abstract}
3D point cloud completion, the task of inferring the complete geometric shape from a partial point cloud, has been attracting attention in the community. For acquiring high-fidelity dense point clouds and avoiding uneven distribution, blurred details, or structural loss of existing methods' results, we propose a novel approach to complete the partial point cloud in two stages. Specifically, in the first stage, the approach predicts a complete but coarse-grained point cloud with a collection of parametric surface elements. Then, in the second stage, it merges the coarse-grained prediction with the input point cloud by a novel sampling algorithm. Our method utilizes a joint loss function to guide the distribution of the points. Extensive experiments verify the effectiveness of our method and demonstrate that it outperforms the existing methods in both the Earth Mover's Distance (EMD) and the Chamfer Distance (CD). 
\end{abstract}

\section{Introduction}
\begin{figure}[t]
  \centering
  \includegraphics[width=\linewidth]{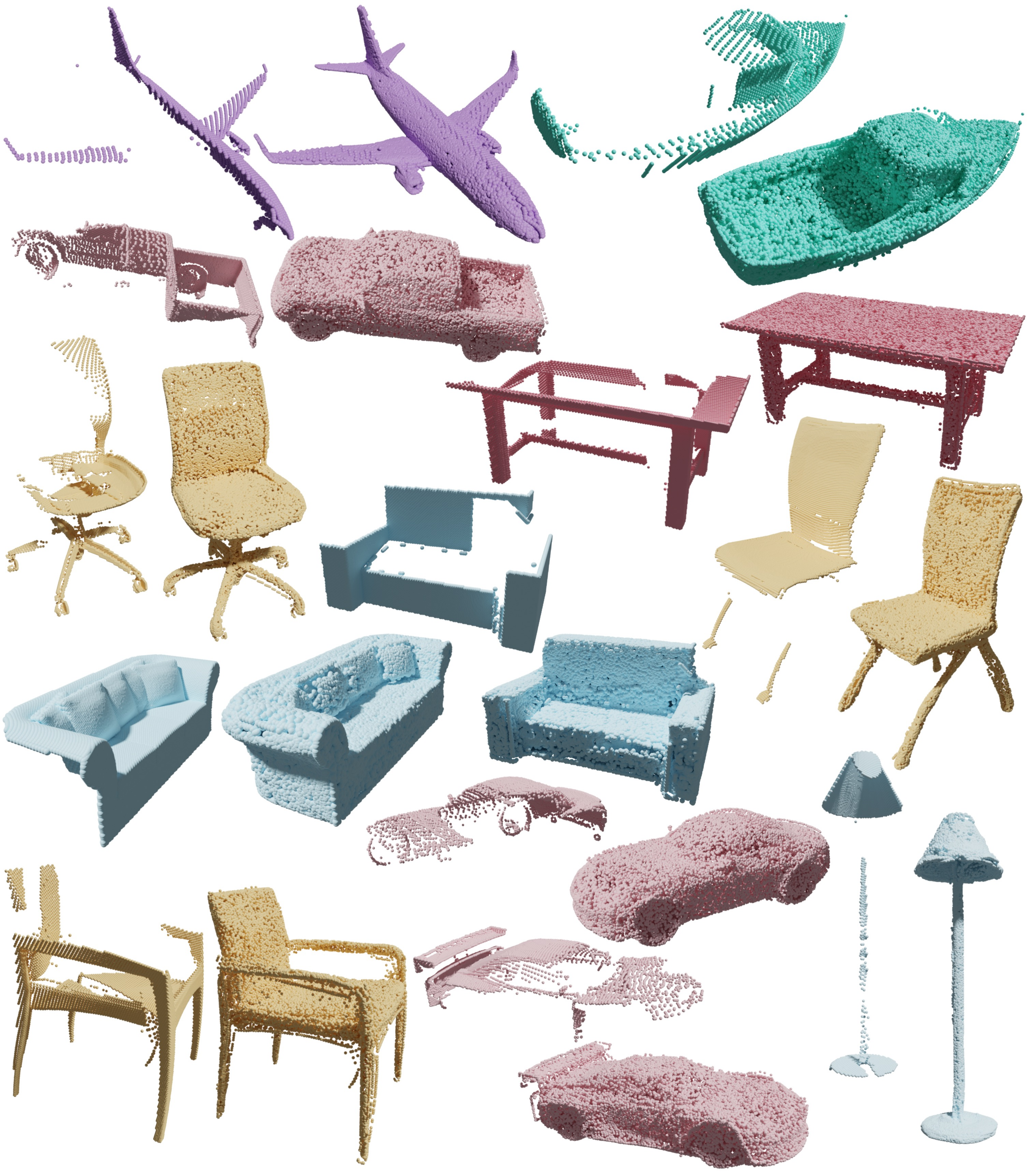}
   \caption{
   Our network predicts realistic structures from partial views and completes the point clouds evenly. Each pair (before and after completion) is visualized in the same color, with 32,768 points after completion.}
   \label{denseresult}
\end{figure}
Acquiring high-fidelity 3D models from real-world scans is challenging, which not only depends on the capability of sensors but also relies on sufficient views for scanning. Based on such restricted raw data, shape completion is required to compensate for the structural loss and enhance the quality, in order to benefit subsequent applications, such as shape classification~\cite{Sarmad_2019_CVPR} and point cloud registration~\cite{yuan2018pcn}.

Existing learning-based methods represent 3D shapes as volumetric grids~\cite{dai2017shape,han2017high,Stutz2018} or view-based projection~\cite{park2017transformation} and then leverage 3D/2D convolution operations. These methods suffer from high computational cost or loss of geometric information. With the advances in deep learning for point cloud analysis and generation, some reasonable works on 3D point cloud completion have been presented~\cite{yuan2018pcn,8658987,Sarmad_2019_CVPR,topnet2019}, which prevents high memory footprints and artifacts caused by discretization. 

However, due to the limited capability of analyzing and generating point clouds, these works sometimes produce distorted results or even fail to preserve some of the actual structures which have been revealed in the input. For example, they may be able to complete the overall shape of a chair, but may neglect the connectors between the chair legs although they appear in the input point cloud. On the other hand, the similarity metric for point cloud comparison plays an important role. The widely used Chamfer Distance (CD) tends to cause uneven density distribution and blurred details~\cite{achlioptas2017learning}. As an alternative, the Earth Mover's Distance (EMD) is more sensitive to details and the density distribution, yet suffers from high computational cost and thus have not been applied to dense point clouds.

To tackle these problems, we propose a novel network which completes the partial point cloud in two stages. In the first stage, we follow the auto-encoder architecture, and utilize a collection of 2-manifold like surface elements, which can be 2D parameterized, to assemble a complete point cloud. In order to prevent surface elements from overlapping, we propose an expansion penalty which motivates each surface element to be concentrated in a local area. Although we can predict a complete point cloud with only such an auto-encoder, the surface generation may be coarse-grained, and the prediction may also neglect some structures within the input. To this end, in the second stage, we combine the coarse-grained prediction with the input point cloud and employ a novel sampling algorithm to obtain an evenly distributed subset point cloud from the combination. A point-wise residual is then learned for the point cloud which enables fine-grained details. We use EMD to compare with the ground truth and utilize an auction algorithm~\cite{bertsekas1992auction} for the EMD approximation, which can be applied to dense point clouds. 

Extensive experiments verify the effectiveness of our novelties. Our method outperforms the existing methods with regard to both EMD and CD. Figure~\ref{denseresult} shows some completion results. The contribution of our work mainly includes:
\begin{itemize}
\item a novel approach for dense point cloud completion, which preserves known structures and generates continuous and smooth details;
\item expansion penalty for preventing overlaps between the surface elements;
\item a novel sampling algorithm for obtaining an evenly distributed subset point cloud;
\item an implementation of the EMD approximation, which can be applied to dense point clouds.
\end{itemize}

\section{Related Work}

\textbf{3D Shape Completion} Conventional methods for 3D shape completion mainly includes geometry-based approaches and example-based approaches. Geometry-based approaches may interpolate smooth surfaces based on existing structures~\cite{davis2002filling,zhao2007robust} or rely on some geometric assumptions, such as symmetry~\cite{thrun2005shape,sipiran2014approximate}. However, interpolation-based methods do not apply to the cases with large-scale incompleteness, and geometric assumptions do not always hold true for real-world 3D data. Example-based approaches~\cite{pauly2005example,sung2015data,shen2012structure} first retrieve some similar models in a large shape database, and then deform and assemble the retrieved models to complete the partial shape. The shape database plays an important role in example-based approaches, making them impractical for completing rare-seen novel shapes.

Learning-based approaches utilize a parametric model (e.g., neural network) to learn a mapping between the partial shape and its completion. Lots of works resort to volumetric grids and leverage 3D convolution networks~\cite{dai2017shape,han2017high,Stutz2018}. 3D shapes can be projected into 2D views and some methods use 2D convolution operations for novel view generation~\cite{tatarchenko2016multi,park2017transformation}. Representing 3D shapes as polygon meshes,~\citeauthor{litany2018deformable} completes the partial human body and face meshes with the help of graph convolution and the reference mesh models. There are also some recent methods exploring continuous implicit filed for representing 3D shape~\cite{chen2018learning,park2019deepsdf}. However, these methods have their own limitations, such as high computational cost, loss of geometric details, and applicability to only certain shape categories.

\textbf{Point Cloud Analysis} Without the loss of geometric information and the artifact from the discretization, point clouds can be a more efficient representation. However, since point clouds are unordered and may have varying densities, deep learning on irregular point clouds faces many challenges and we cannot apply traditional convolution on point clouds directly. PointNet~\cite{qi2017pointnet} uses symmetric functions to aggregate information from individual points, followed by some improvements on local feature learning~\cite{qi2017pointnet++,shen2018mining}. Some methods project point clouds to regular structures, which allows traditional convolution~\cite{su2018splatnet,tatarchenko2018tangent}. By constructing graphs for point clouds, some approaches employ graph-based analysis~\cite{hu2018semantic,landrieu2018large,wang2018dynamic}. There are also lots of works exploring specialized convolution operation for point clouds~\cite{jiang2018pointsift,li2018pointcnn,liu2019relation}.

\textbf{Point Cloud Generation} Decoding point clouds from latent features has not been fully explored.~\citeauthor{fan2017point} generate point cloud coordinates using a fully-connected branch and a 2D deconvolution branch. FoldingNet~\cite{yang2018foldingnet} deforms a 2D plane into a 3D shape, which favors continuous and smooth structures. Combining the merits of the fully-connected layer and FoldingNet, PCN~\cite{yuan2018pcn} proposes a coarse-to-fine point cloud generator. AlasNet~\cite{groueix2018papier} further represents 3D shape as a collection of parametric surface elements and learns the mappings from the 2D square to 3D surface elements, which enables generating complex shapes. 


\begin{figure*}[t]
  \centering
  \includegraphics[width=\linewidth]{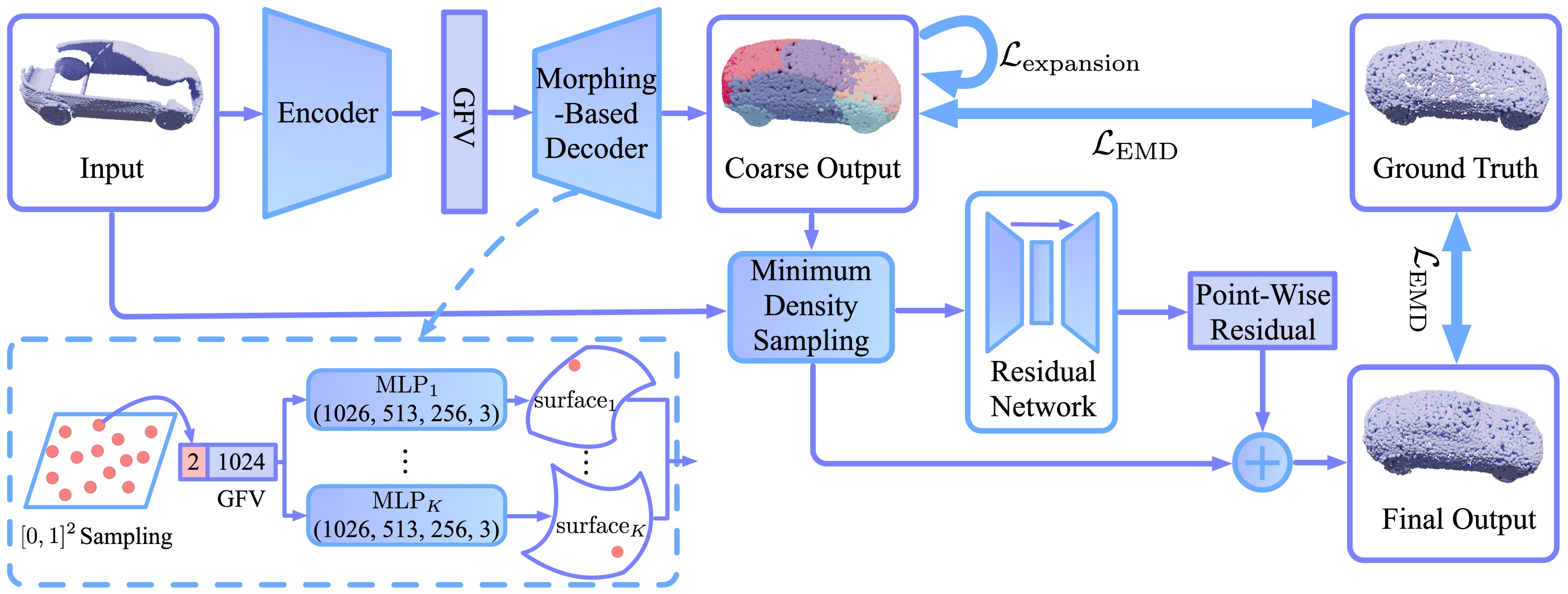}
  \caption{Architecture of our approach. ``GFV'' denotes the generalized feature vector, $\mathcal{L}_{\mathrm{expansion}}$ and $\mathcal{L}_{\mathrm{EMD}}$ denotes the expansion penalty and Earth Mover's Distance respectively. ``$[0,1]^2$ Sampling'' denotes sampling 2D points on a unit square. The morphing-based decoder morphs the unit squares into a collection of surface elements, which are assembled into the coarse output. The minimum density sampling outputs an evenly distributed subset point cloud.}
  \label{framework}
\end{figure*}

\section{Approach}

Given a point cloud lying on the partial surface of an object, our approach is expected to predict a point cloud indicating the complete shape of the object. The output point cloud should be dense enough and evenly distributed so that it can capture the details of the shape. Our approach leverages supervised learning and is trained end-to-end. 

As shown in Figure~\ref{framework}, our approach takes a partial point cloud as input and completes it in two stages. In the first stage, the auto-encoder predicts a complete point cloud by morphing the unit squares into a collection of surface elements. The expansion penalty is proposed to prevent the overlaps between the surface elements. In the second stage, we merge the coarse output with the input point cloud. Through a special sampling algorithm, we obtain an evenly distributed subset point cloud from the combination, and then feed it into a residual network for point-wise residual prediction. By adding the residual, our approach outputs the final point cloud. Unlike many existing approaches, we employ EMD for dense point cloud comparison.

\subsection{Morphing-Based Prediction} 
In the first stage, we hope to predict a point cloud, which captures the overall shape of the object, with an auto-encoder. For efficiency, the encoder is designed following the idea of PointNet~\cite{qi2017pointnet}, though we could use other networks for feature extracting as well. Inspired by the AtlasNet~\cite{groueix2018papier}, we then feed the extracted features into a morphing-based decoder for predicting continuous and smooth shapes.

As shown in the bottom left of Figure~\ref {framework}, the decoder employs $K$ (16 in experiments) surface elements to form a complex shape. Each surface element is expected to focus on a local area which is relatively simple, making the generation of local surfaces easier. For each element, the decoder learns a mapping from the unit square $[0,1]^2$ to the 3D surface using a multilayer perceptron (MLP), which mimics the morphing of a 2D square into a 3D surface. In each forward pass, we randomly sample $N$ (512 in experiments) points in the unit square. The encoded feature vector, which describes the prediction, is then concatenated with the sampled point coordinates, before passing them as input to the $K$ MLPs. Each sampled 2D point will be mapped to $K$ 3D points lying on the $K$ different surface elements. As a result, each forward pass outputs $KN$ (8,192 in experiments) points describing the predicted shape. Since the MLPs learn continuous mappings from 2D to 3D, the decoder can generate smooth surfaces by dense sampling on 2D. While many approaches, like the fully-connected layer, output fix-sized discrete point coordinates, our approach can combine the results from multiple forward passes to generate point clouds with arbitrary resolution. For instance, Figure~\ref{denseresult} shows dense point clouds generated by 4 forward passes.

\begin{figure}[t]
  \centering
  \includegraphics[width=\linewidth]{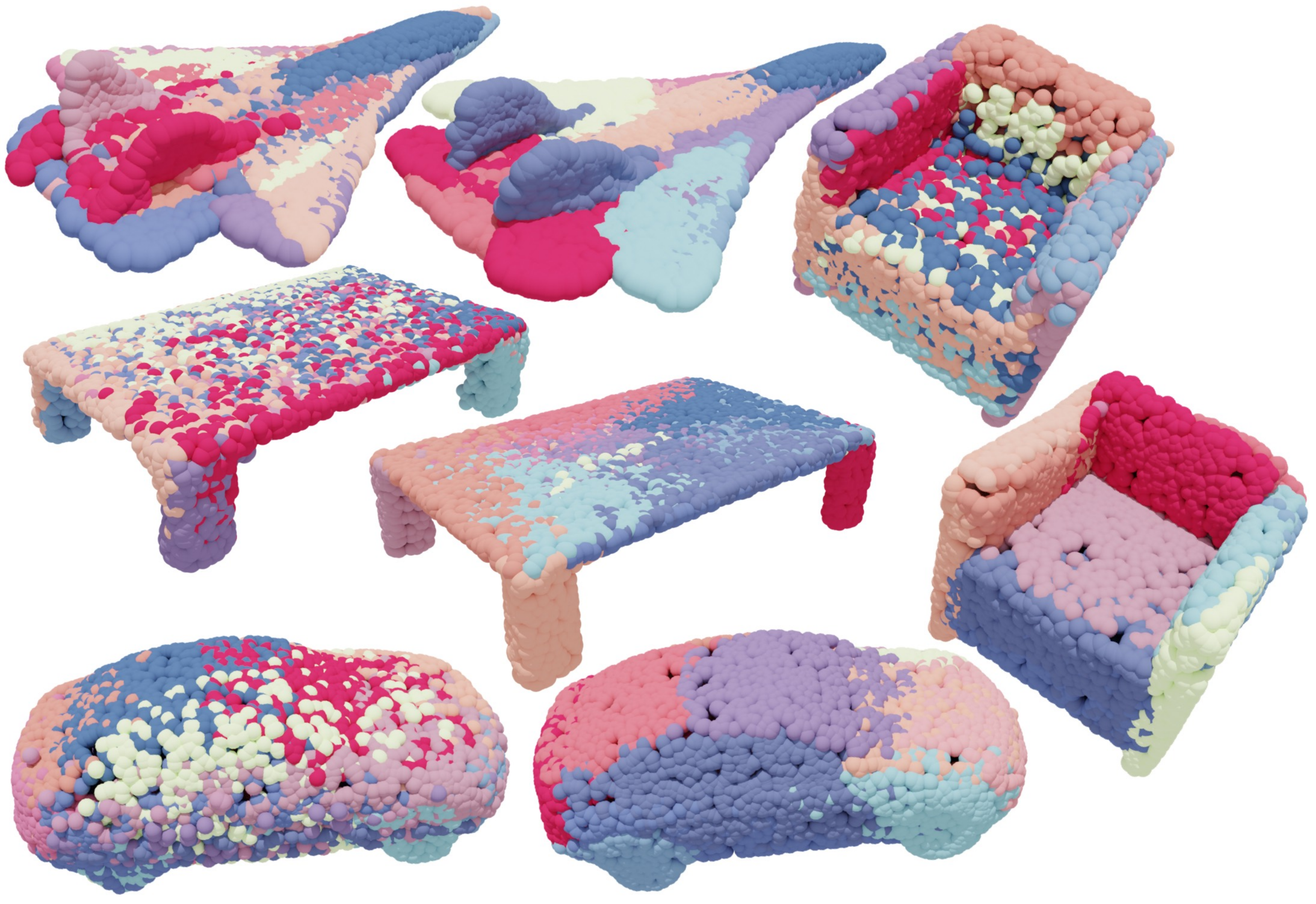}
  \caption{The figure shows pairs of point clouds. Points from the same surface elements are in the same color. In each pair, the left one is the result without the expansion penalty, where different surface elements tend to mix with each other. The right one is the result with the expansion penalty.}
  \label{fig:mst_result}
\end{figure}

\begin{figure}[t]
  \centering
  \includegraphics[width=1\linewidth]{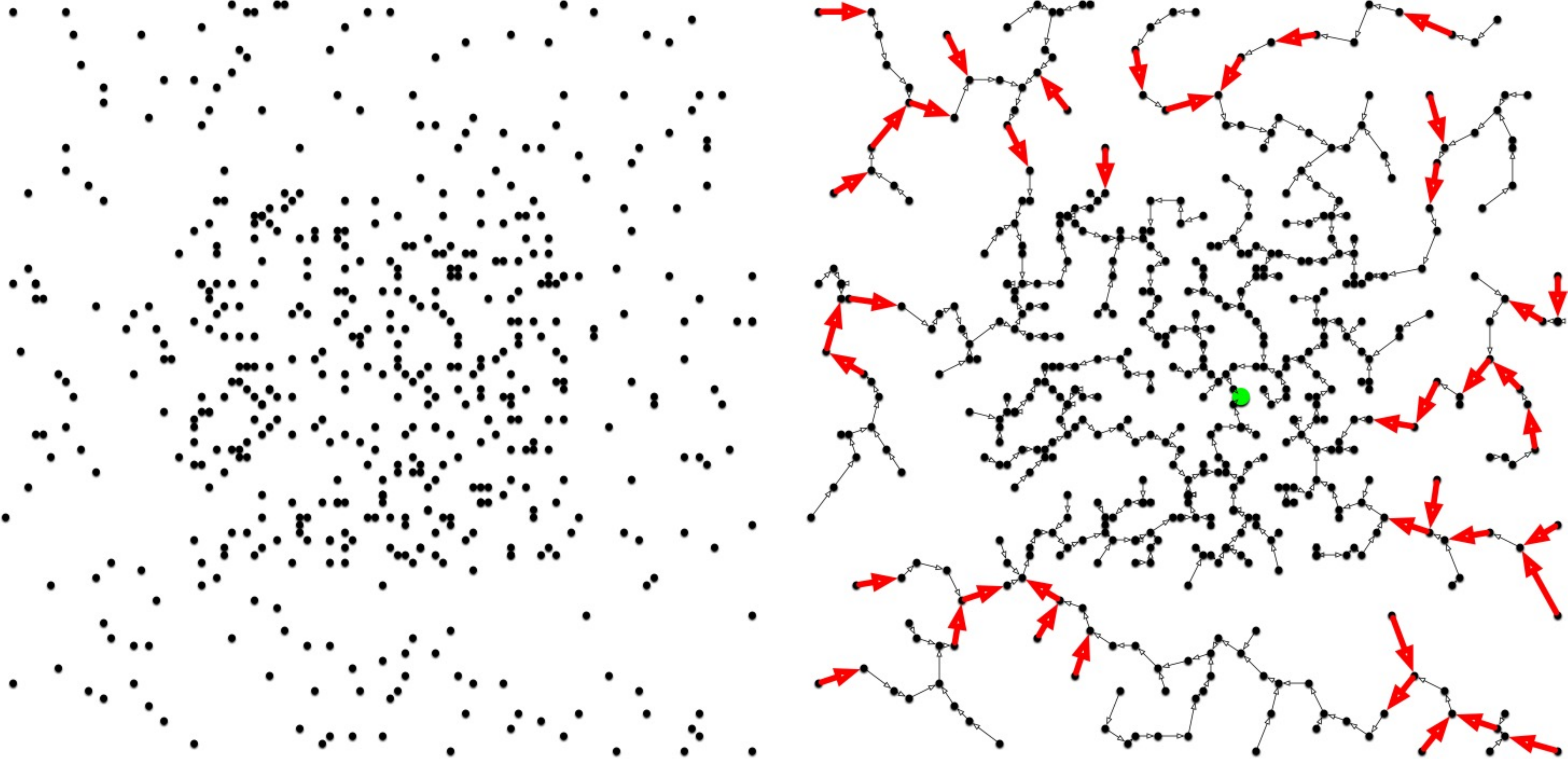} 
  \caption{The left figure shows the point cloud of a surface element (2D for clear presentation). The right figure shows the directed minimum spanning tree, where the green point indicates the root. For all red edges $(u,v)$, $u$ is motivated to shrink toward $v$.}
  \label{mst_construction}
\end{figure}
Although we will use similarity metrics (e.g., EMD) to guide the union of the MLPs cover the whole shape of the prediction, the MLPs in the AtlasNet~\cite{groueix2018papier} are not explicitly prevented from generating the same area of space, which may result in overlaps between the surface elements. Figure~\ref{fig:mst_result} shows such an example, in the left point cloud of each pair, points from different surface elements tend to mix with each other. The overlaps may lead to the uneven density distribution of the point cloud. It may also cause surface elements to expand and cover larger areas, which makes it more difficult to morph the 2D square and capture local details.

To this end, we propose an \textbf{expansion penalty}, which serves as a regularizer for surface elements. It encourages each surface element to be compact and concentrated in a local area. Specifically, in each forward pass, we regard the generated 3D points from each MLP as a vertex set and construct a minimum spanning tree $\mathcal{T}_i$~\cite{prim1957shortest} for each of them based on the Euclidean distances between the points. We then choose the middle vertex of $\mathcal{T}_i$'s diameter (i.e., the simple path containing the most vertices) as the root vertex and direct $\mathcal{T}_i$ by making all its edges point toward the root. Figure~\ref{mst_construction} shows an example of the construction. This way, we have $K$ directed minimum spanning trees which describe the distribution of points from each MLP. As shown in Figure~\ref{mst_construction}, edges with longer length (i.e., the distance between the pair of points) suggest more sparsely distributed points which tend to mix with points from other MLPs. The expansion penalty thus makes those points shrink along the edges toward the more compact areas. It can be formulated as: 
{\small
$$
\mathcal { L } _ { \mathrm {expansion} }\!=\! \frac{1}{KN}\! \sum _ {1 \leq i \leq K} \sum _ { ( u , v ) \in \mathcal{T}_i} \!\!\mathbbm{ 1 }\!\left\{ \operatorname { dis } ( u , v )\!\geq\!\lambda l _ { i } \right\}\!\operatorname { dis } ( u , v )
$$
}
where $\operatorname { dis }(u, v)$ denotes the Euclidean distance between vertex $u$ and vertex $v$, $l_i = (\sum_{( u , v ) \in \mathcal{T}_i}\operatorname { dis }(u, v))/(N - 1)$ denotes the average length of edges in $\mathcal{T}_i$, and $\mathbbm{ 1 }$ is the indicator function filtering edges whose length are shorter than $\lambda l_i$ ($\lambda$ is 1.5 in experiments). The expansion penalty is differentiable almost everywhere, since the constructed spanning tree is invariant under the infinitesimal movement of the points. For each directed edge $(u,v) \in \mathcal{T}_i$, whose length is longer than $\lambda l _ { i }$, we only give $u$ a gradient in the backward passes. So that $u$ is motivated to shrink toward $v$, making a more compact surface element. 

As shown in Figure~\ref{fig:mst_result}, thanks to the expansion penalty, the overlaps between the surface elements are mitigated. The MLPs divide the whole shape into $K$ parts and each MLP covers a local part. The partition even corresponds to the semantic parts of the object, which shows the potential for the downstream semantic applications. 

Although there may be more intuitive methods to motivate each surface element to be concentrated (e.g., the distances to the mean point), we find them over-constrain the shape of each element. The key idea of our spanning-tree based method is that we only want to penalize those points which are sparsely distributed (e.g., those sparsely distributed points on the boundary of each surface element) instead of all the points. Our expansion penalty thus allows each surface element to generate more flexible shapes.

\subsection{Merging and Refining} 
With the morphing-based auto-encoder, we can generate a smooth point cloud predicting the overall shape. However, due to the limited capabilities, the auto-encoder may neglect some structures, which have been revealed in the input point cloud. Also, the fixed-sized surface elements are not flexible enough for fine-grained local details. Therefore, we merge the coarse output from the auto-encoder with the input and then learn a point-wise residual for the combination.  

\begin{figure}[t]
  \centering
  \includegraphics[width=\linewidth]{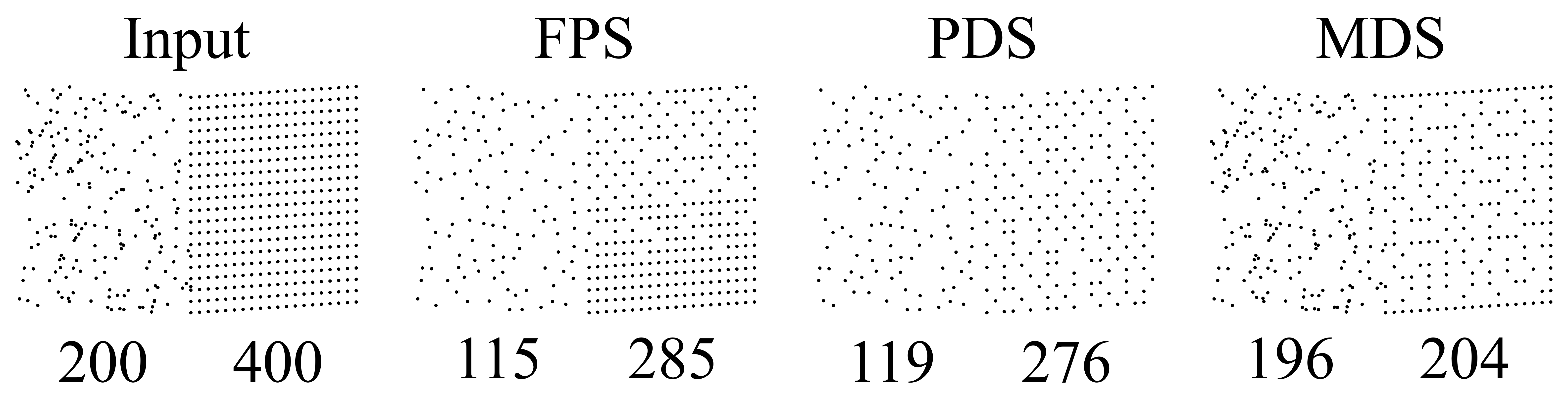}
  \caption{The figure shows the results of sampling 400 points from 600 points. The global distribution of MDS's result is more uniform than that of FPS and PDS.}
  \label{sampling}
\end{figure}

Since the density of the two point clouds may be different and there may be overlapping between them, the merged point cloud is probably unevenly distributed. We thus hope to sample a subset point cloud, which has a uniform distribution, from the combination. Existing sampling algorithms for point clouds, such as the farthest point sampling (FPS) and Poisson disk sampling (PDS)~\cite{wei2008parallel}, cannot guarantee the global density distribution of the results. Figure~\ref{sampling} shows such an example: the input point cloud consists of two parts, with the right part being twice as dense as the left part. The sampling results of FPS and PDS are unevenly distributed. Although the area is the same, the number of points on the right side is much larger than the number of points on the left side. Inspired by the point cloud uniformization algorithm using graph Laplacian~\cite{luo2018uniformization}, we employ the summation of Gaussian weights to estimate the ``density'' of a point and thus propose a novel sampling algorithm namely \textbf{minimum density sampling (MDS)}. We denote the $i$th sampled point as $p_i$ and the set of first $i$ sampled points as $\mathrm{P}_i=\{p_j| 1\leq j\leq i\}$. Unlike FPS returning the farthest point from $\mathrm{P}_{i-1}$ as $p_i$, in each iteration, MDS returns a point that has the minimum ``density'':
{\small
$$
p_i =  \argmin_{x\notin \mathrm{P}_{i-1}}  \sum _ { p_j\in \mathrm{P}_{i-1}} \exp ( -  \| x - p_j \|^2 / (2\sigma^2)  )
$$ 
}
where the parameter $\sigma$ is a positive quantity, which corresponds to the size of the neighborhood considered. As shown in Figure~\ref{sampling}, MDS outputs a subset point cloud whose global distribution is more uniform than that of FPS and PDS.

Taking the evenly distributed subset point cloud as input, we then learn a point-wise residual for refinement, which enables the generation of fine-grained structures. Since the points from the input are more reliable, in addition to the channels of the point coordinates, we add another binary channel to the input to distinguish the source of each point, where ``0'' stands for the input point cloud and ``1'' for the coarse output. The architecture of the residual network resembles PointNet~\cite{qi2017pointnet}, which consumes a point cloud and outputs a three-channel residual. We output the final point cloud after adding the residual point by point.

\subsection{Similarity Metric} 
\begin{figure}[t]
  \centering
  \includegraphics[width=0.8\linewidth]{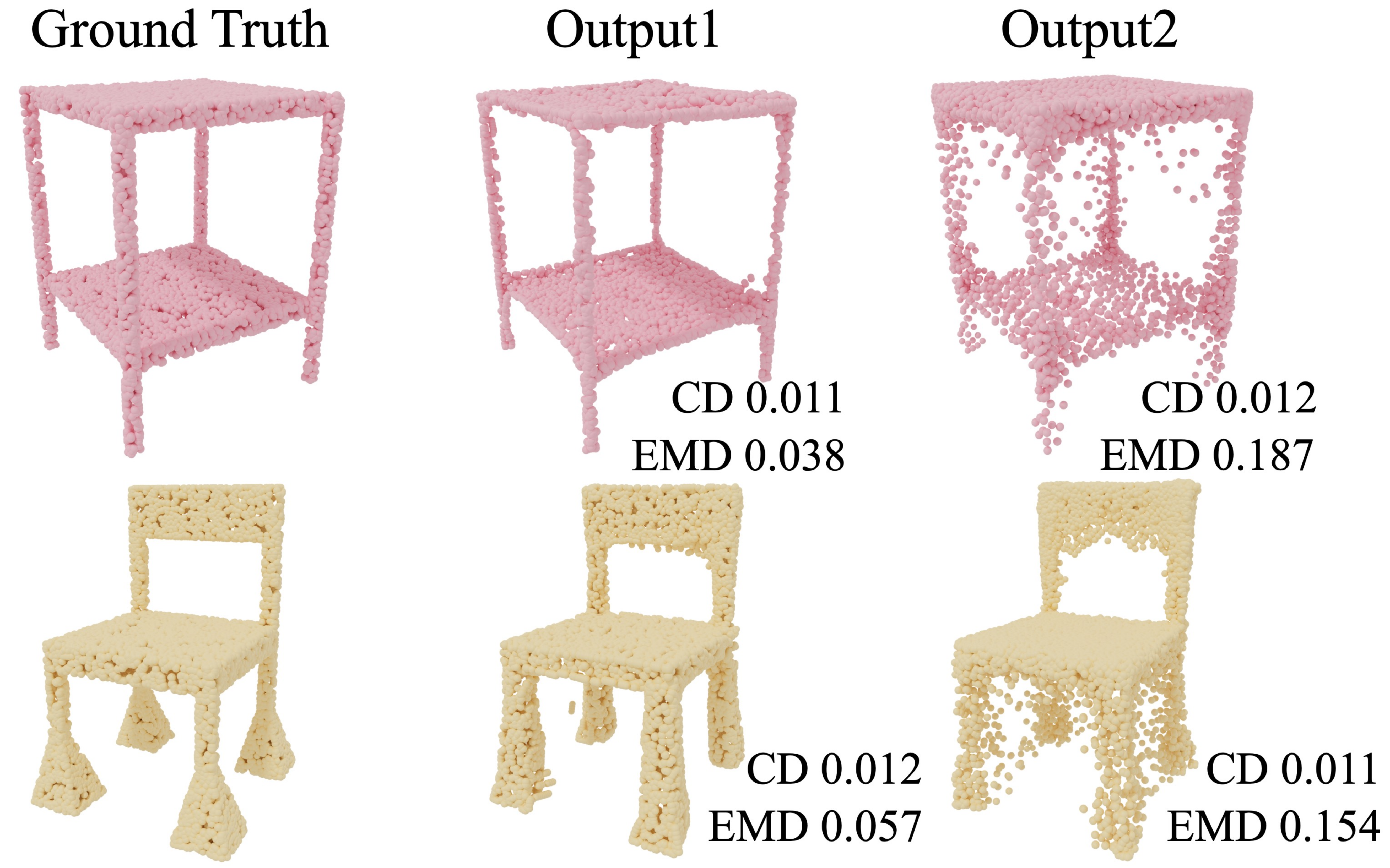}
  \caption{The figure shows the completion results of two different methods. The EMD is more reliable to distinguish the visual quality of the results.}
  \label{fig:CDEMD}
\end{figure}

One of the challenges in point cloud completion is the comparison with the ground truth. Existing similarity metrics mainly include the Chamfer Distance (CD) and the Earth Mover's Distance (EMD)~\cite{fan2017point}. For two point clouds $S_1$ and $S_2$, CD measures the mean distance between each point in one point cloud to its nearest neighbor in the other point cloud:
{\small
$$
\mathcal{L}_{\mathrm{CD}}(S_{1},\!S_{2})\!\!=\!\!\frac{1}{2}\!\left( \frac{1}{|\!S_1\!|}\sum_{x\in S_1}\!\min_{y\in S_2}\!\|x\!-\!y\|\!+\!\frac{1} {|\!S_2\!|}\!\sum_{ y \in S_2}\!\min_{x \in S_1}\!\|x\!-\!y\|\right)
$$
}
EMD is only defined when $S_1$ and $S_2$ have the same size: 
{\small
$$
\mathcal{L}_ { \mathrm { EMD } } \left( S _ { 1 } , S _ { 2 } \right) = \min _ { \phi : S _ { 1 } \rightarrow S _ { 2 } } \frac { 1 } { |S _ { 1 }| } \sum _ { x \in S _ { 1 } } \| x - \phi ( x ) \| _ { 2 }
$$
}
where $\phi$ is a bijection. Most existing works employ CD as a loss function, since it's more efficient to compute. However, CD is blind to some visual inferiority~\cite{achlioptas2017learning}. Figure~\ref{fig:CDEMD} shows such an example. In the outputs of the second method, points tend to over-populate in those locations where most objects of that category have mass (e.g., table tops for tables), and the details of the changeable parts are always blurred. However, it's hard for CD to penalize this type of cheat since one of its summands can be very small and the other one can also be not so large. 

By solving the linear assignment problem, EMD forces the output to have the same density distribution as the ground truth and is thus more discriminative to the local details and the density distribution. Most existing methods employ an implementation for EMD approximation which needs $O(n^2)$ memory footprints ($n$ denotes the number of points) and cannot be applied to dense point clouds due to the memory bottleneck. Therefore, many existing works~\cite{yuan2018pcn,achlioptas2017learning,wang20193dn} use EMD only for point clouds with about 2,000 points, which is not sufficient to capture many details.

Inspired by the auction algorithm~\cite{bertsekas1992auction}, a constant approximation for the linear assignment problem, we implement an approximation for EMD, which only needs $O(n)$ memory. It can thus be applied to dense point clouds for comparing more details. Specifically, the algorithm treats the points from the two point clouds as persons and objects respectively and finds an economic equilibrium by proceeding the bidding phases and assignment phases iteratively. The algorithm terminates in finite iterations and outputs an assignment of points whose the mean distance is within $\epsilon$ of being optimal. Here, $\epsilon$ is a parameter which balances the error rate and the speed of convergence. In training processes, to accelerate the calculation of the loss function, we fix the number of iterations and assign the remaining points greedily. The whole procedure of our implementation is parallelized enabling deep learning with GPUs. 

In fact, there are many other variants of the transportation distances~\cite{cuturi2013sinkhorn,solomon2015convolutional}, and it's promising to explore whether they can be applied to comparing two point clouds in an efficient and effective manner.

Our joint loss function $\mathcal{L}$ can thus be calculated as: 
{\small
$$
\mathcal{L}\!=\!\mathcal{L}_ { \mathrm { EMD } }\!\left(\mathrm{S}_{\mathrm{coarse}}, \mathrm{S}_{\mathrm{gt}}\right)+\alpha \mathcal{L}_{\mathrm{expansion}} + \beta \mathcal{L}_ { \mathrm { EMD } }\!\left(\mathrm{S}_{\mathrm{final}}, \mathrm{S}_{\mathrm{gt}}\right)
$$
}
where $\mathrm{S}_{\mathrm{coarse}}$ denotes the coarse output, $\mathrm{S}_{\mathrm{final}}$ denotes the final output, and $\mathrm{S}_{\mathrm{gt}}$ denotes the ground truth. $\alpha$ and $\beta$ are weighting factors ($\alpha$ is $0.1$ and $\beta$ is $1.0$ in experiments). The EMD terms and the expansion penalty work in opposite directions. The former motivates the point cloud to cover the whole shape of the object, while the latter serves as a regularizer which encourages each surface elements to shrink. Their mutual restraint allows each surface element to be centralized in a local area while the union of the elements is as close as possible to the ground truth.

\section{Experiments}


\subsection{Data Generation and Model Training}

\begin{figure*}[t]
  \centering
  \includegraphics[width=\linewidth]{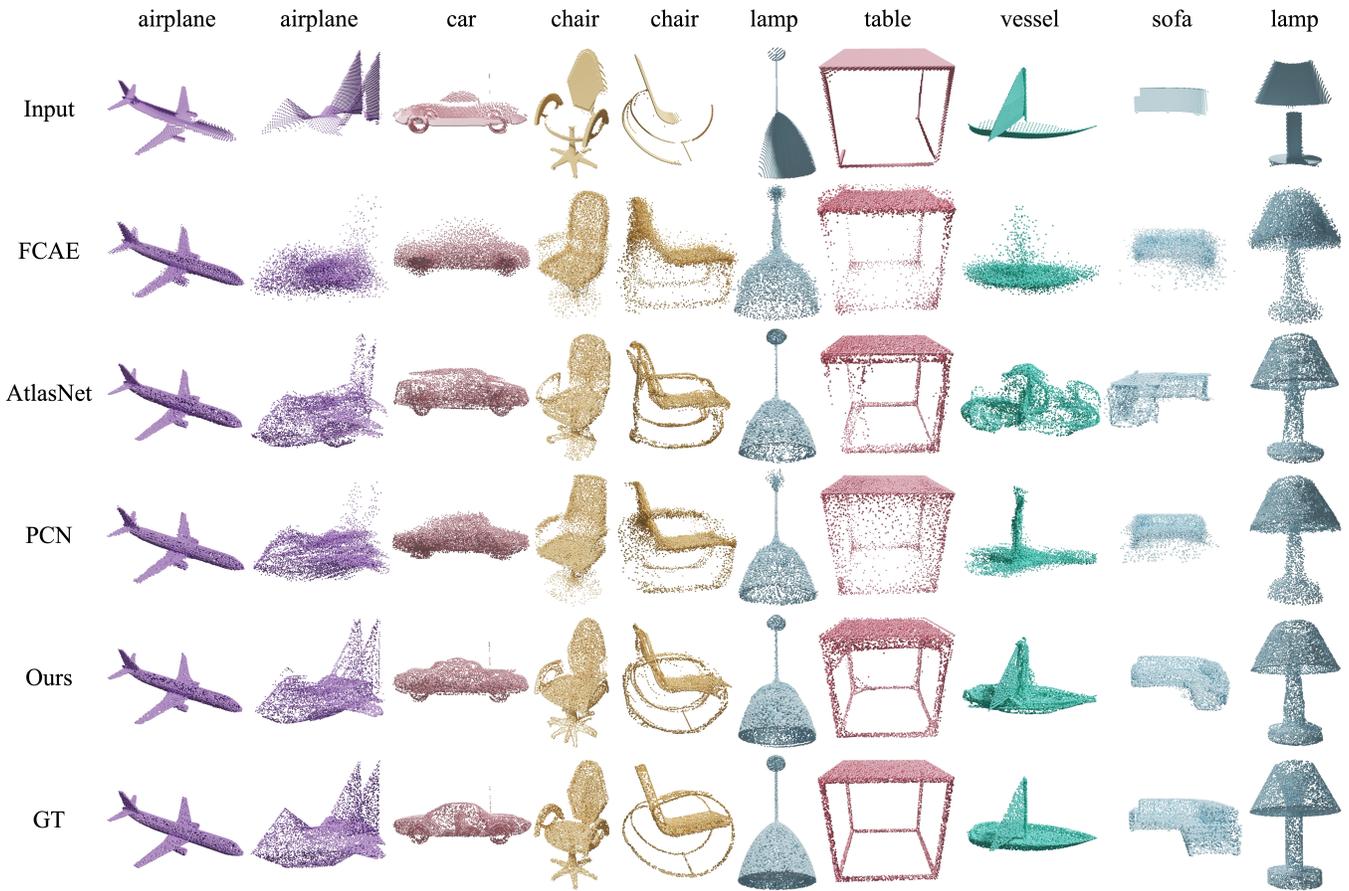}
  \caption{The figure shows the completion results of different methods. Each output point cloud consists of $8\text{,}192$ points.}
  \label{fig:comparison}
\end{figure*}

We evaluate our methods on the ShapeNet dataset~\cite{Shapenet}. We choose 30,974 synthetic CAD models from the dataset, which cover eight categories: table, chair, car, airplane, sofa, lamp, vessel, and cabinet. For a fair comparison, the train/test split is the same as in PCN~\cite{yuan2018pcn}. We generate 50 pairs of partial and complete point clouds for each of the CAD model, resulting in 30,974 $\times$ 50 pairs of point clouds for training and test. Specifically, the CAD models are normalized and located at the origin. For each of them, we uniformly sample 8,192 points on the surface, which form the complete point cloud. We then randomly sample 50 camera poses on the unit sphere and lift the 2.5D captured images into 3D partial point clouds, which mimics obtaining 3D raw data from different views in the real-world applications. After denoising, for convenience of training, the partial point clouds are unified into the size of $5\text{,}000$ points by randomly dropping or replicating points.

We trained our models on 8 Nvidia GPUs for $2.3 \times 10^5$ iterations (i.e., 25 epochs) with a batch size of 160. The initial learning rate is 1e-3 and is decayed by 0.1 per 10 epochs. Adam is used as the optimizer. All activation functions are ReLU except for the final tanh layers producing the point coordinates and residuals. The models are trained end-to-end for all object categories.

\begin{table}[t] 
\caption{Quantitative comparison between our methods and existing methods. For both EMD and CD, lower is better.}  
\small
\begin{subtable}{\linewidth}
\centering
 \setlength{\tabcolsep}{1.5pt}
    \begin{tabular}{c|cccccccc|c}
    \hline
    {\small methods}    & {\scriptsize vessel} & {\scriptsize cabinet} & {\scriptsize table} & {\scriptsize airplane} & {\scriptsize car} & {\scriptsize chair} & {\scriptsize sofa} & {\scriptsize lamp} & {\small average} \\
    \hline
    {\small Oracle} & 0.93  & 1.44  & 1.21  & 0.69  & 1.25  & 1.17  & 1.28  & 0.91  & 1.11  \\
    \hline
    {\small FCAE}  & 7.22 &  11.20 &   7.77 &  4.10 &  7.00 &  7.64 &  7.00 &  14.64 &   8.32   \\
    {\small AtlasNet} & 8.11  & 8.91  & 5.07  & 3.27  & 4.20  & 5.03  & 6.97  & 10.71  & 6.53  \\
    {\small PCN}   & 6.56  & 8.79  & 6.84  & 3.44  & 4.44  & 6.89  & 6.28  & 15.45  & 7.34  \\
    {\small Ours}  & \textbf{3.83} & \textbf{4.16} & \textbf{3.66} & \textbf{2.18} & \textbf{3.28} & \textbf{3.63} & \textbf{3.47} & \textbf{6.04} & \textbf{3.78} \\
    \hline
    \end{tabular}%
    \caption{$\mathrm{EMD}\times 100$}
\end{subtable}

\begin{subtable}{\linewidth}
\centering       
\setlength{\tabcolsep}{1.5pt}
    \begin{tabular}{c|cccccccc|c}
    \hline
    {\small methods}    & {\scriptsize vessel} & {\scriptsize cabinet} & {\scriptsize table} & {\scriptsize airplane} & {\scriptsize car} & {\scriptsize chair} & {\scriptsize sofa} & {\scriptsize lamp} & {\small average} \\
    \hline
    Oracle & 0.42  & 0.72  & 0.55  & 0.28  & 0.63  & 0.50  & 0.60  & 0.35  & 0.50  \\
    \hline
    FCAE  & 1.33  & 1.40  & 1.16  & 0.70  & 1.10  & 1.29  & 1.37  & 1.72  & 1.26  \\
    AtlasNet & 2.30  & 2.49  & 1.46  & 0.85  & 1.42  & 1.58  & 2.67  & 1.82  & 1.82  \\
    PCN   & 1.23  & 1.35  & 1.14  & 0.66  & 1.10  & 1.41  & 1.36  & 1.46  & 1.21  \\
    Ours  & 1.17  & 1.37  & 1.15  & 0.60  & 1.11  & 1.16  & 1.31  & 1.30  & 1.14  \\
    Ours-CD & \textbf{0.99} & \textbf{1.19} & \textbf{0.96} & \textbf{0.56} & \textbf{1.03} & \textbf{1.02} & \textbf{1.16} & \textbf{1.07} & \textbf{1.00} \\
    \hline
    \end{tabular}%
    \caption{$\mathrm{CD}\times 100$}
\end{subtable}  
\label{table:quantitative}
\end{table}  

\subsection{Comparison with Existing Methods}

We compare our approach to the following methods. \textbf{FCAE} uses an intuitive auto-encoder where the encoder follows PointNet~\cite{qi2017pointnet} and the decoder is a fully connected layer generating the coordinates of 8,192 points directly. CD is used as the loss function. \textbf{AtlasNet}~\cite{groueix2018papier} employs the similar encoder but generates the point cloud with a set of parametric surface elements. Since AtlasNet outputs 2,500 points per forward pass, we combine the generated points of 4 different passes and then randomly sample 8,192 points from the combination. \textbf{PCN}~\cite{yuan2018pcn} completes the partial point cloud through an auto-encoder as well. It utilizes a stacked version of PointNet~\cite{qi2017pointnet} as the encoder and generates point clouds in a coarse-to-fine fashion. We randomly sample 8,192 points from the output for comparison. The \textbf{Oracle} method randomly samples $8\text{,}192$ points on the complete surface of the object. The reported distances between the randomly sampled points and the ground truth point cloud provide an estimation of the upper bound of the performance. \textbf{Ours-CD} is the same as our method except for replacing EMD with CD.

The quantitative results are reported in Table~\ref{table:quantitative}. 
It can be seen that our methods outperform existing methods with regard to both EMD and CD. As mentioned earlier, compared to CD, EMD is more discriminative and convincing for point cloud comparison. Our method achieves the lowest EMD in all object categories and the average EMD is only $57.9\%$ and $51.5\%$ to that of AtlasNet and PCN, which demonstrates the superiority of our method. The EMD differs in various object categories suggesting the difficulties of completing different categories varies. Specifically, there are more airplanes and cars in the dataset and their structures are relatively simple and stable, making them easier to complete. In contrast, various lamps are relatively isolated in the dataset, which are more difficult to complete. As for CD, the discrepancies among different methods and object categories are relatively small. Although ``Ours'' uses EMD as the loss function while the existing methods use CD, ``Ours'' outperforms them with regard to CD and ``Ours-CD'' achieves lower results. However, there is still a huge gap between the results of completion methods and the ``Oracle'', which indicates the completion task is arduous.

Figure~\ref{fig:comparison} shows qualitative results. All methods work well with simple cases, such as the first airplane, but the discrepancy of the methods appears in complex cases. Specifically, FCAE and PCN tend to generate blurred details which may be the results of using the fully connected layers to output coordinates directly. In contrast, our method predicts more realistic structures and generates continuous and smooth details. Also, our method employs EMD as the loss function, which guarantees the even distribution of the points, while existing methods are more likely to overpopulate points in some parts. It can also be seen that our method can preserve the known structures, while other methods always distort or even neglect the structures revealed in the input. Moreover, unlike FCAE and PCN outputting the fix-sized point clouds, our method is capable of generating dense point cloud with arbitrary resolution. Figure~\ref{denseresult} shows such an example.

\subsection{Ablation Study}

\begin{table}[t]
  \centering
  \small
  \setlength{\tabcolsep}{1.5pt}
  \caption{The ablated versions of our method. ``$\checkmark$'' indicates including that component while ``$\times$'' indicates not.}
    \begin{tabular}{cccccc}
    \hline
     methods & $\mathcal{L}_{\mathrm{expansion}}$ &  merging  & refining & CD/EMD \\
    \hline
     w/o $\mathcal{L}_{\mathrm{expansion}}$ (A)     &   $\times$   &   MDS   & \checkmark   & EMD \\
    w/o merging (B)     &   \checkmark   &  $\times$    & $\times$   & EMD \\
    w/o MDS (C)     &   \checkmark   &  FPS   & \checkmark   & EMD \\
    w/o refining (D)     &   \checkmark   &  MDS    & $\times$   & EMD \\
    Ours-CD (E)     &   \checkmark   &  MDS    & \checkmark   & CD \\
    Ours  &   \checkmark   &  MDS    & \checkmark   & EMD \\
    \hline
    \end{tabular}%
  \label{tab:ablation}%
\end{table}%

\begin{table}[t] 
\caption{Quantitative comparison between our method and the ablated versions. For descriptions of the methods, see Table~\ref{tab:ablation} and the text.  For both EMD and CD, lower is better.}  
\label{tab:ablation_result}
\centering  
\small
\begin{subtable}{\linewidth}
\centering       
\setlength{\tabcolsep}{1.5pt}
    \begin{tabular}{c|cccccccc|c}
    \hline
    {\small methods}    & {\scriptsize vessel} & {\scriptsize cabinet} & {\scriptsize table} & {\scriptsize airplane} & {\scriptsize car} & {\scriptsize chair} & {\scriptsize sofa} & {\scriptsize lamp} & {\small average} \\
    \hline
    A     & 3.94  & 4.33  & 3.85  & 2.23  & 3.47  & 3.78  & 3.59  & 6.08  & 3.91  \\
    B     & 4.18  & 4.37  & 4.08  & 2.39  & 3.46  & 3.89  & 3.75  & 6.51  & 4.08  \\
    C     & 4.30  & 5.30  & 4.24  & 2.59  & 4.01  & 4.41  & 4.18  & 6.38  & 4.43  \\
    D & 3.93 & 4.32 & 3.73 & 2.38 & 3.41 & 3.73 & 3.64 & 6.02 & 3.89 \\ 
    E     & 5.44  & 6.81  & 4.52  & 3.01  & 4.39  & 5.44  & 5.62  & 8.93  & 5.52  \\
    Ours     & \textbf{3.83} & \textbf{4.16} & \textbf{3.66} & \textbf{2.18} & \textbf{3.28} & \textbf{3.63} & \textbf{3.47} & \textbf{6.04} & \textbf{3.78} \\
    \hline
    \end{tabular}%
    \caption{$\mathrm{EMD}\times 100$}
\end{subtable}  

\begin{subtable}{\linewidth}
\centering    
    \setlength{\tabcolsep}{1.5pt}
    \begin{tabular}{c|cccccccc|c}
    \hline
    {\small methods}    & {\scriptsize vessel} & {\scriptsize cabinet} & {\scriptsize table} & {\scriptsize airplane} & {\scriptsize car} & {\scriptsize chair} & {\scriptsize sofa} & {\scriptsize lamp} & {\small average} \\
    \hline
    A     & 1.20  & 1.46  & 1.22  & 0.62  & 1.15  & 1.23  & 1.38  & 1.37  & 1.20  \\
    B     & 1.36  & 1.48  & 1.29  & 0.70  & 1.19  & 1.30  & 1.45  & 1.59  & 1.29  \\
    C     & 1.09  & 1.38  & 1.12  & 0.58  & 1.11  & 1.10  & 1.27  & 1.23  & 1.11  \\
    D & 1.24 & 1.44 & 1.21 & 0.64 & 1.15 & 1.23 & 1.40 & 1.39 & 1.21 \\ 
    \textbf{E}     & \textbf{0.99} & \textbf{1.19} & \textbf{0.96} & \textbf{0.56} & \textbf{1.03} & \textbf{1.02} & \textbf{1.16} & \textbf{1.07} & \textbf{1.00} \\
	Ours     & 1.17  & 1.37  & 1.15  & 0.60  & 1.11  & 1.16  & 1.31  & 1.30  & 1.14  \\
    \hline
    \end{tabular}%
    \caption{$\mathrm{CD}\times 100$}
\end{subtable}  
\end{table}  

\begin{table}[t]
  \centering
  \caption{The surface area for the ground truth, and the sum of the areas of the surface elements for the two methods. The numbers are the averages over all object categories.}
    \begin{tabular}{ccc}
    \hline
    ground truth & w/o $\mathcal{L}_{\mathrm{expansion}}$ & Ours  \\
    \hline
    1.11   & 2.41 & 1.56  \\
    \hline
    \end{tabular}%
  \label{tab:area}%
\end{table}%

We compare our method to some ablated versions which are described in Table~\ref{tab:ablation}. With regard to the~\textbf{expansion penalty}, by remeshing the point clouds with the Ball-Pivoting Algorithm~\cite{817351}, we compute the sum of the areas of the surface elements. The results are shown in Table~\ref{tab:area}, which suggests that, without the expansion penalty, each surface element tends to cover a significantly larger area. Figure~\ref{fig:mst_result} demonstrates that the expansion penalty can motivate each surface element to be centralized in a local area and prevent surface elements from overlapping. As shown in Table~\ref{tab:ablation_result}, version A, without the expansion penalty, typically produces point clouds with larger EMD and CD. With regard to the \textbf{merging operation}, merging the coarse output with the input point cloud can preserve some known structures resulting in a more reliable output. Version B, without the merging operation and the subsequent refining operation, typically produces point clouds with larger EMD and CD. Moreover, we use MDS to obtain an evenly distributed subset point cloud. Replacing the MDS with the FPS, version C cannot guarantee the even distribution of the points, which causes a significantly larger EMD. However, since the FPS may preserve more points from the reliable input, version C produces point clouds with smaller CD. With regard to the \textbf{refining operation}, the point-wise residual enables the generation of fine-grained details. Without the refining operation, version D tends to produce point clouds with larger EMD and CD. With regard to the \textbf{similarity metric}, we implement an approximation which allows calculating EMD for dense point clouds. Version E replaces EMD with CD and produces solutions with larger EMD and smaller CD.
 
\section{Conclusion}
We have presented a novel approach for point cloud completion, which completes the partial point cloud in two stages. With the expansion penalty, we can effectively control the distribution of the points. The novel sampling operation enables us to preserve the known structures. Moreover, we have discussed the similarity metrics and implemented an efficient approximation for EMD. Extensive experiments demonstrate that our approach predicts more realistic structures and generates dense point clouds evenly.
\bibliographystyle{aaai}
{
\fontsize{8.1pt}{9.1pt} \selectfont
\bibliography{bib}
}
\end{document}